\documentclass[runningheads]{llncs}

\usepackage{eccv}

\usepackage{eccvabbrv}

\usepackage{graphicx}
\usepackage{booktabs}
\usepackage{tikz}
\usepackage{multirow}
\newcommand\blfootnote[1]{%
  \begingroup
  \renewcommand\thefootnote{}\footnote{#1}%
  \addtocounter{footnote}{-1}%
  \endgroup
}

\usepackage[accsupp]{axessibility}  

\usepackage[pagebackref,breaklinks,colorlinks,citecolor=eccvblue]{hyperref}

\usepackage{orcidlink}

\begin{document}

\title{SynFundus-1M: A High-quality Million-scale Synthetic fundus images Dataset with Fifteen Types of Annotation} 


\author{Fangxin Shang\inst{1} \and
        Jie Fu\inst{2} \and
        Yehui Yang\inst{1*} \and
        Haifeng Huang\inst{1} \and
        Junwei Liu\inst{1} \and
        Lei Ma\inst{3}
}

\authorrunning{Shang et al.}
\titlerunning{SynFundus-1M}

\institute{
    Healthcare Group, Baidu Inc., Beijing, China \and
    Hong Kong University of Science and Technology, Hong Kong, China \and
    Peking University, Beijing, China
    \url{https://github.com/parap1uie-s/SynFundus-1M}
}

\maketitle

\begin{abstract}
Large-scale public datasets with high-quality annotations are rarely available for intelligent medical imaging research, due to data privacy concerns and the cost of annotations.
In this paper, we release SynFundus-1M, a high-quality synthetic dataset containing over \textbf{One million} fundus images in terms of \textbf{eleven disease types}.
Furthermore, we deliberately assign four readability labels to the key regions of the fundus images.
To the best of our knowledge, SynFundus-1M is currently the largest fundus dataset with the most sophisticated annotations. 
Leveraging over 1.3 million private authentic fundus images from various scenarios, we trained a powerful Denoising Diffusion Probabilistic Model, named SynFundus-Generator. 
The released SynFundus-1M are generated by SynFundus-Generator under predefined conditions. 
To demonstrate the value of SynFundus-1M, extensive experiments are designed in terms of the following aspect: 
1) Authenticity of the images: we randomly blend the synthetic images with authentic fundus images, and find that experienced annotators can hardly distinguish the synthetic images from authentic ones. Moreover, we show that the disease-related vision features (e.g. lesions) are well simulated in the synthetic images.
2) Effectiveness for down-stream fine-tuning and pretraining: we demonstrate that retinal disease diagnosis models of either convolutional neural networks (CNN) or Vision Transformer (ViT) architectures can benefit from SynFundus-1M, and compared to the datasets commonly used for pretraining, models trained on SynFundus-1M not only achieve superior performance but also demonstrate faster convergence on various downstream tasks. 
SynFundus-1M is already public available for the open-source community.

\keywords{Fundus image\and Synthetic images\and Pretraining}
\end{abstract}

\section{Introduction}
\label{sec:intro}

\blfootnote{* Corresponding author, e-mail: yangyehuisw@126.com}

\begin{table}
\centering
    \resizebox{0.95\linewidth}{!}{
\begin{tabular}{ccccccc}
\hline
\multirow{2}{*}{Dataset} & \multirow{2}{*}{Available images} & \multicolumn{5}{c}{Annotations} \\ \cline{3-7} 
 &  & Glaucoma & \begin{tabular}[c]{@{}c@{}}Diabetic\\ retinopathy\end{tabular} & \begin{tabular}[c]{@{}c@{}}Pathological\\ myopia\end{tabular} & Other diseases & Readability \\ \hline
AIROGS \cite{devente23airogs} & 101,442 & \checkmark &  &  & 0 & 0 \\
EyePACS \cite{cuadros2009eyepacs} & 88,702 &  & \checkmark &  & 0 & 0\\
APTOS \cite{aptos2019-blindness-detection} & 5,590 &  & \checkmark &  & 0 & 0 \\
Messidor-2 \cite{decenciere2014feedback} & 1,744 &  & \checkmark &  & 0 & 0 \\
PALM \cite{fu2019palm} & 1,200 &  &  & \checkmark & 0 & 0 \\
IDRiD \cite{porwal2020idrid} & 1,113 &  & \checkmark &  & 0 & 0 \\
REFUGE2 \cite{fang2022refuge2} & 2,000 & \checkmark &  &  & 0 & 0 \\ \hline
SynFundus-1M & \textbf{1,000,018} & \checkmark & \checkmark & \checkmark & 8 & 4 \\ \hline
\end{tabular}}
\caption{\label{tab:datasets}Comparative overview of data volume and annotations across fundus imaging datasets, highlighting SynFundus-1M as the dataset with the most sophisticated annotations.}
\end{table}

Fundus imaging is one of the most important bases for enhancing early detection and accurate treatment of eye diseases, significantly improving patient outcomes and transforming eye care efficiency.
The past years have witnessed tremendous efforts in introducing deep learning methods for the automatic analysis of fundus images \cite{gulshan2016development,yang2017lesion,yang2020residual,yang2021robust,zhou2020encoding,zhou2022generalizable}.

It is widely recognized that the performance of deep learning models is closely tied to the quantity and quality of training data \cite{ji2023exploring,boecking2022making,xu2018towards}, and there exist several popular open-source fundus datasets. 
However, the number of images or the annotation categories of the existing datasets are limited. 
As depicted in Table~\ref{tab:datasets}, AIROGS \cite{devente23airogs}, the currently largest fundus image dataset, is exclusively annotated for glaucoma classification. 
Similarly, the commonly used EyePACS \cite{cuadros2009eyepacs} only contains annotation of diabetic retinopathy grades. 

In the field of medical imaging, acquiring favorable train data can be challenging due to privacy concerns and annotation cost \cite{lotan2020medical, singh2020current}. 
Therefore, despite the availability of numerous open-source medical imaging datasets, most of them in a dilemma of balancing the quantity and variety of high-quality annotations.

Due to the difficulty of obtaining promising authentic medical images, some researchers have resorted to using synthetic data to enhance model performance~\cite{wu2023datasetdm,azizi2023synthetic, khader2022medical}. 
Recently, diffusion models \cite{ho2020denoising,azizi2023synthetic} have been demonstrated to surpass traditional generative adversarial networks (GANs) related approaches in various applications.\cite{dhariwal2021diffusion, muller2022diffusion}. 

Inspired by the diffusion models, we train a Denoising Diffusion Probabilistic Model (DDPM), named \textit{SynFundus-Generator}, with a unique and private dataset comprising 1.3 million authentic fundus images. 
Using the SynFundus-Generator, we generate a collection of over one million synthetic fundus images, named \textit{SynFundus-1M} dataset, with 15 types of annotation (11 disease labels and 4 image readability labels). Compared to the open-source datasets in Table~\ref{tab:datasets}, the released SynFundus-1M offers not only a large scale of fundus images, but also boasts a broader spectrum of complete annotations w.r.t. various disease and image readability. 

Extensive experiments prove that our synthetic images can hardly be distinguished from the authentic ones even by four experienced annotators, and the synthetic disease-related visual features are also authentic to be indistinguishable. As described in Section \ref{sec:downstream_exp} and \ref{sec:pretrain}, the fundus image analysis models achieve performance improvements by utilizing SynFundus-1M for both fine-tuning and pre-training purposes. Researchers can also directly use this dataset or further refine the labels to suit their specific needs.

To contribute to the community, we release SynFundus-1M to pave the way for advancements in retinal disease diagnostics. We hope that this study will fuel further breakthroughs and broaden the application scope of the fundus imaging analysis, as well as providing a new approach to open-source high-quality data while ensuring data privacy.

\section{Material and Methods}
\subsection{The Training Data}

We take much efforts to craft the training data of SynFundus-Generator, which is the key to the quality of the released SynFundus-1M. The training data rests upon the following pivotal pillars: 
\begin{enumerate}
    \item \textbf{The number and variety of authentic images}: The scale and variety of a dataset significantly influence the efficacy of generative models. We are privy to a private dataset comprising over \textbf{1.3 million} authentic fundus images. These images encompass a range of retinal diseases, captured under different conditions in clinic scenarios, including health examination scene, outpatient scene, hospitalization scene etc.

    \item \textbf{The reliable annotations}: Given the scale of training images, the cost of sophisticated disease annotation can be prohibitive. 
    Our team has been engaged in fundus AI research for more than six years. We've developed a AI-assisted system for fundus analysis, which obtained the National Medical Products Administration (NMPA) class III certificate\footnote{The NMPA Class III Medical Devices represent the most stringent certification system for medical devices in China, requiring rigorous control over both safety and effectiveness.}. 
    Over these years, we've accumulated tens of thousands of annotated fundus images encompassing 11 disease categories and 4 image readability labels. Leveraging these high-quality annotations, we've developed numerous models for both disease and image readability classification.
    These models have undergone rigorous testing in real-world scenarios, demonstrating a sensitivity and specificity exceeding 90\% during practical evaluation, thus validating their robust performance in real-life applications.
    We use these models as AI-diagnose platform to fill up the missing labels of private training images. Therefore, The training data of SynFundus-Generator encompasses a comprehensive set of 15 labels, covering both fundus diseases and image readabilities.
\end{enumerate}
 

\subsection{The SynFundus-Generator}

\begin{figure}[!ht]
\centering
\includegraphics[width=1.0\columnwidth,trim=30 180 15 80,clip]{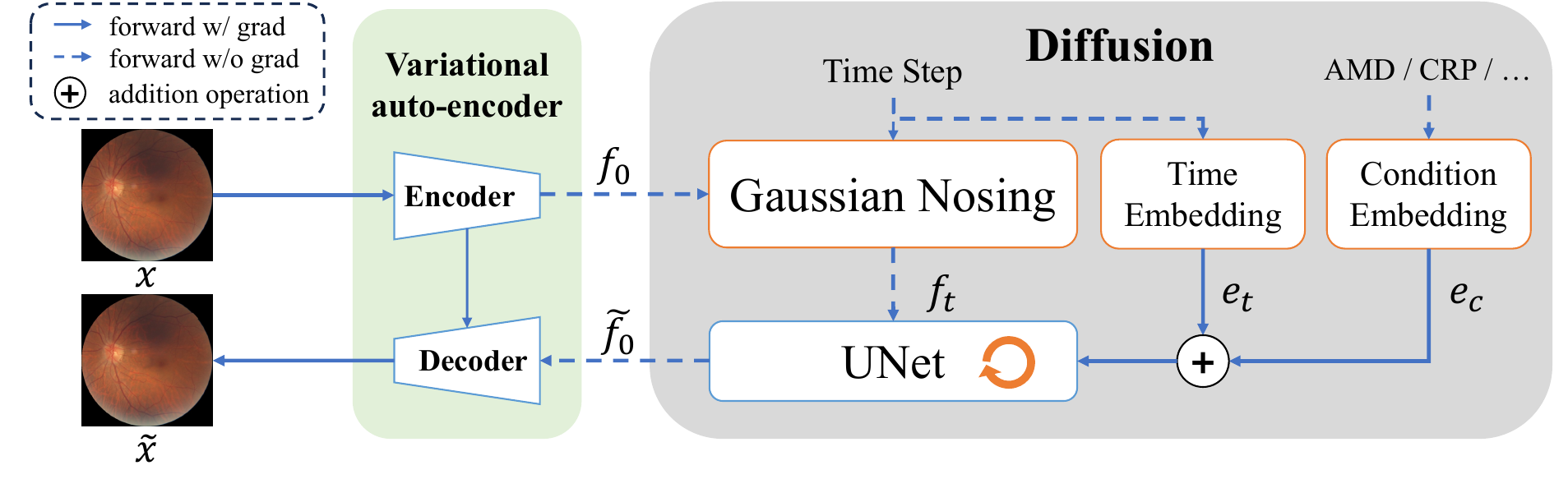}
\caption{\label{fig:illustration}Illustration of the SynFundus-Generator. 
Generate conditions are embedded to the noise estimation procedure. 
The orange arrow \raisebox{-.3em}{\includegraphics[width=1em]{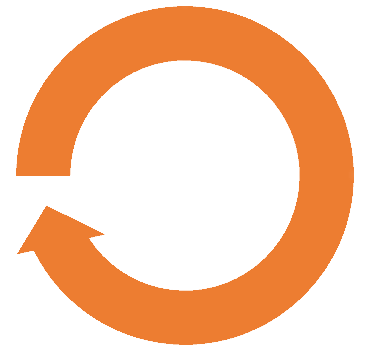}} represents the noise estimator iteratively loops through the estimation noise and generation process.}
\end{figure}

In this research, we trained a variant of the latent denoising diffusion probabilistic model, termed MedFusion \cite{muller2022diffusion}, as the \textit{SynFundus-Generator}. The training procedure of SynFundus-Generator consists of two stages, as illustrated in Figure \ref{fig:illustration}.

\textbf{VAE stage}: Firstly, a variational auto-encoder (VAE) model is trained to minimize the gap between the input RGB image $x$ and the output decode result $\tilde{x}$. 
The trained VAE encode images from RGB space into an 8-times compressed latent space (64x64 for the 512x512 input space). 
This procedure aims to construct an information bottleneck and enforce the latent code to extract the most relevant visual representation of the fundus content. 

\textbf{Diffusion stage}: Secondly, the latent code $f_0$ generated by VAE encoder with up to $T=1000$ steps Gaussian noise, named $f_T$, is fed into diffusion model. 
A U-Net model is used to estimate the noise added at step $t\in[0,T]$, then the latent code with noise $\tilde{f_t}$ can be denoised under specifically indicated conditions at $t$. 
The diffusion model (U-Net) is trained with the conditions (disease, readability and $t$) to minimize the mean squared error (MSE) loss between the noise added in current step and the prediction. 

This denoise procedure is iterative at generation to obtain the final denoised latent code $\tilde{f_0}$ from a sample of random noise. 
The denoised $\tilde{f_0}$ is fed into the VAE decoder to generate the synthetic image. 
Standard image augmentation techniques are used in training VAE, including random vertical, horizontal flipping and image rotation, and only random crop techniques apply for the diffusion model.

\subsection{The Construction of SynFundus-1M} \label{sec:synfundus_gen}

\begin{table}
\centering
\resizebox{\linewidth}{!}{
\begin{tabular}{ccll} 
\hline
Category & Abbreviation & \multicolumn{1}{c}{Explain} & \multicolumn{1}{c}{Annotation Type} \\ 
\hline
\multirow{11}{*}{Disease} & DR & Diabetic Retinopathy & \begin{tabular}[c]{@{}l@{}}\textbf{Enum[0,1,2,3,4].}\\ 0 (non-diabetic eye disease), \\ 1 (mild non-proliferative), \\ 2 (moderate non-proliferative), \\ 3 (severe non-proliferative),\\ 4 (proliferative).\end{tabular} \\ 
\cline{4-4}
 & AMD & Age-related Macular Degeneration & \multirow{10}{*}{\begin{tabular}[c]{@{}l@{}}\textbf{Bool.}\\~\\ True (Positive for the disease).\\ False (Negative for the disease).\end{tabular}} \\
 & AON & Anomalies of the Optic Nerve &  \\
 & CRP & Choroidal Retinal Pathology &  \\
 & DM & Degenerative Myopia &  \\
 & DME & Diabetic Macular Edema &  \\
 & EM & Epimacular Membrane &  \\
 & GC & Glaucoma &  \\
 & HtR & Hypertensive Retinopathy, &  \\
 & PM & Pathological Myopia &  \\
 & RVO & Retinal Vein Occlusion &  \\ 
\hline
\multirow{4}{*}{Readability} & Readable & Is readable fundus & \multirow{4}{*}{\begin{tabular}[c]{@{}l@{}}\textbf{Bool.}\\ True (Positive for the readability).\\ False (Negative for the readability).\end{tabular}} \\
 & RO & Readability of Optical disc &  \\
 & RR & Readability of Retinal region &  \\ 
 & RM & Readability of Macular &  \\ 
\hline
\end{tabular}}
\caption{\label{tab:dataset_annotation}The scope and meaning of the annotations in SynFundus-1M.}
\end{table}

The released SynFundus-1M dataset is generated by SynFundus-Generator. 
A comprehensive description of the annotations is provided in Table~\ref{tab:dataset_annotation}. 
There are eleven disease and four readability types which covers the most common fundus diseases in clinical. 

\begin{figure}[!ht]
\centering
\includegraphics[width=1.0\columnwidth,trim=0 0 0 0,clip]{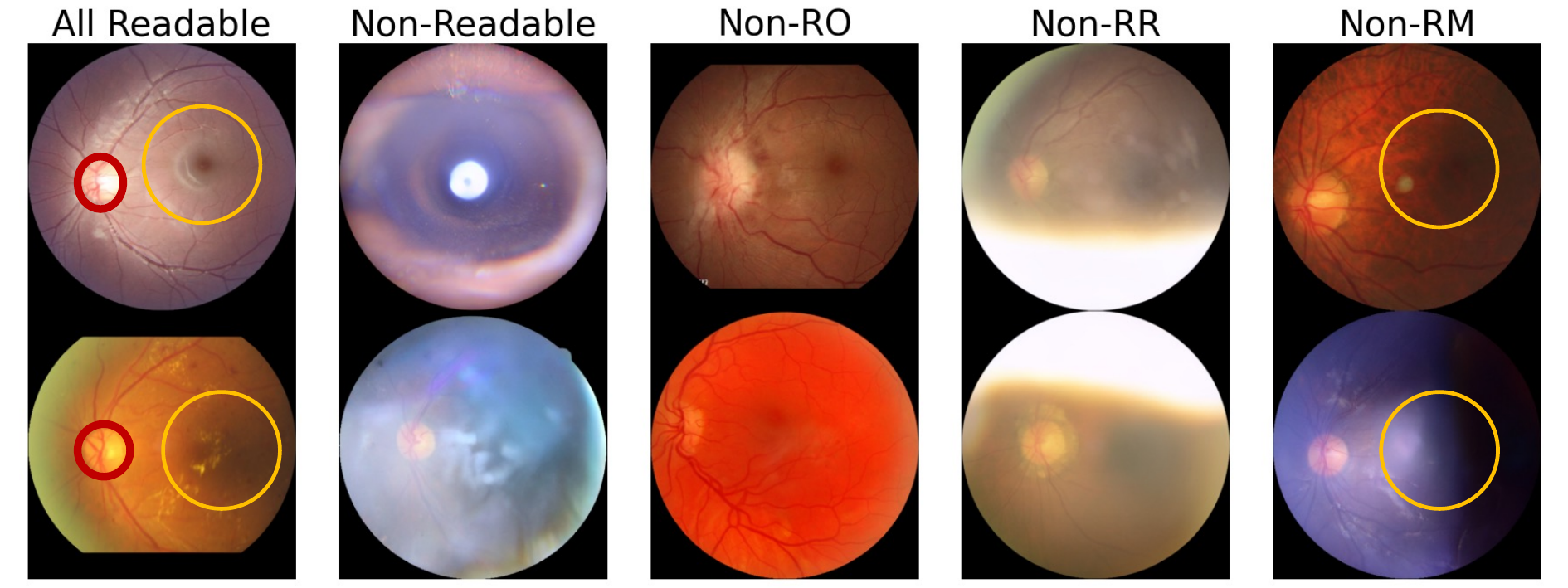}
\caption{\label{fig:readability}An illustration showcasing the various of readability within the SynFundus-1M dataset, where the first column is reabable images, and the remaining columns are various non-readable samples. 
In order to facilitate explanation and comparison, the optical disc are outlined by a red circle, the macular region is marked by a yellow circle.}
\end{figure}

Figure \ref{fig:readability} shows different readability types of images from the SynFundus-1M dataset, with readable images in first column and various unreadable images in remaining columns. 
In some cases, artifacts, poor focus, underexposure, or overexposure in crucial fundus image areas like the optical disc and macular may affect the readability of the key regions, thereby hindering confident clinical conclusions. 
The diagnostic models are anticipated to demonstrate robust performance even some parts of fundus images are in low-readability.
Therefore, we provide four readability labels in SynFundus-1M, defining as follows: 

\textbf{Is readable fundus}: this label is set to \textit{False} if the given image can not provide any useful information for clinical judgement. 

\textbf{Optical disc readability}: Optical disc is a important region in the fundus, which related with multiple disease such as glaucoma and anomalies of the optic nerve. An object detection model was developed to identify the Region of Interest (ROI) of the optic disc within fundus images. Once the ROI is successfully detected, the corresponding object is considered readable for further analysis.

\textbf{Retinal region readability}: The retinal region is defined as the region of the given image except optical disc region. Similar with the judgment readable fundus, if one cannot make the decision in terms of the abnormality from the retinal region, the label of retinal region readability is \textit{False}. 

\textbf{Macular region readability}: Macular is part of the retina at the back of the eye, which is responsible for our central vision, most of our colour vision and the fine detail of what we see. The macular region consists of the fovea and the perifoveal areas. If over one third of macular region are affected by the noise such as artifacts or exposed problems, the macular region readability label is set to \textit{False}. 

For the label distribution in SynFundus-1M, we first examined the distribution and proportions of diseases within the private dataset. 
To increase the ratio of disease-positive samples in SynFundus-1M, the number of samples with exclusively negative labels is manually reduced. 
This analysis served as a foundation for setting the generation conditions for the SynFundus-Generator to generate the dataset. 
All synthetic images are automatically annotated by the AI-diagnose platform to construct the SynFundus Annotation. 
The final annotation distribution is as listed in Table \ref{tab:analysis_label}.

Besides the diseases annotation, most generated samples are high-readability fundus with readable optic and retinal region. 
A small number of low-readability images are retained to improve the robustness for authentic image acquisition scenes. 

The generation of the images for SynFundus-1M entailed approximately 120 hours of computation on A100 GPUs, with an additional 550 hours on V100 GPUs dedicated to annotating the images through the AI-diagnose platform. 

\begin{table}
\centering
\resizebox{0.7\linewidth}{!}{
\begin{tabular}{ccccccccc}
\hline
Label & AMD & AON & CRP & DM & DME & DR & EM & GC \\ \hline
0 & 864224 & 948850 & 870625 & 645573 & 908718 & 804442 & 981295 & 893540 \\
1 & 135794 & 51168 & 129393 & 354445 & 91300 & 56 & 18723 & 106478 \\
2 & - & - & - & - & - & 97882 & \multicolumn{1}{l}{-} & \multicolumn{1}{l}{-} \\
3 & - & - & - & - & - & 51971 & \multicolumn{1}{l}{-} & \multicolumn{1}{l}{-} \\
4 & - & - & - & - & - & 45667 & \multicolumn{1}{l}{-} & \multicolumn{1}{l}{-} \\ \hline
Label & HtR & PM & RVO & \multicolumn{2}{c}{Readable} & RO & RR & RM \\
0 & 965883 & 880964 & 982579 & \multicolumn{2}{c}{84397} & 36352 & 3356 & 28057 \\
1 & 34135 & 119054 & 17439 & \multicolumn{2}{c}{915621} & 963666 & 996662 & 971961 \\ \hline
\end{tabular}}
\caption{\label{tab:analysis_label}Annotation distribution of SynFundus-1M. 
The image readability equals 1 indicates that all regions within the image are readable. 
RO is the abbreviation of readability of optical disc, RR indicates the readability of retinal region and RM indicates the readability of macular.}
\end{table}

\section{Experiments}
\subsection{Authenticity of the SynFundus-1M}

\begin{figure}[!ht]
\centering
\includegraphics[width=1.0\columnwidth,trim=0 0 0 0,clip]{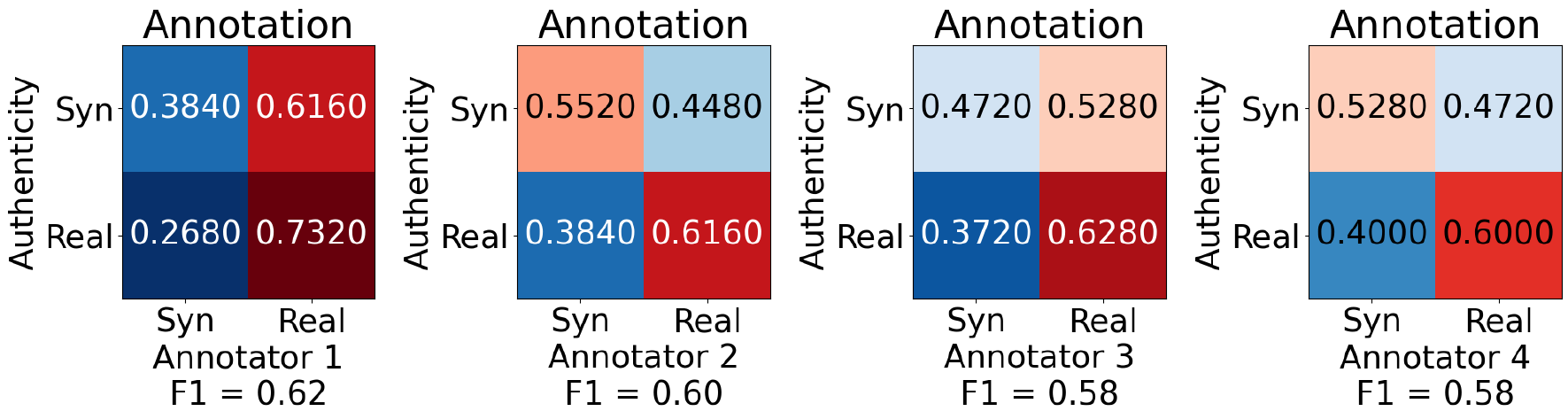}
\caption{\label{fig:human_cm}Confusion matrix displaying the ability of four annotators to discern the authenticity of fundus images. 
Category \textit{Syn} indicates synthetic images, while category \textit{Real} denotes authentic images. 
Metrics nearing 0.5 reflect the annotators' challenges in distinguishing between images, underscoring the authenticity of SynFundus-1M.}
\end{figure}

In this section, we assess the authenticity of synthetic images by both human annotators and the commonly used Fréchet Inception Distance (FID) metric.

For human evaluation, we construct a blending set comprising 250 synthetic images from the SynFundus-1M dataset and 250 authentic images from the EyePACS dataset.
We ensured an equitable distribution across disease categories, aiming for an approximately even number of samples per disease. 
Four experienced annotators, with over five years' experience in developing fundus image algorithms, are requested to independently authenticate the images in the blending set. 

Figure \ref{fig:human_cm} presents the confusion matrices in distinguish the authenticity of the images in the blending set. We can see that the F1-scores of all the four annotators are around 0.6, which indicate that the experienced annotators only achieved a bit higher F1-score than randomly classification. In other words, the experienced specialists struggle to distinguish synthetic images from authentic counterparts.

\begin{figure}[!ht]
\centering
\includegraphics[width=1.0\linewidth,trim=0 0 0 0,clip]{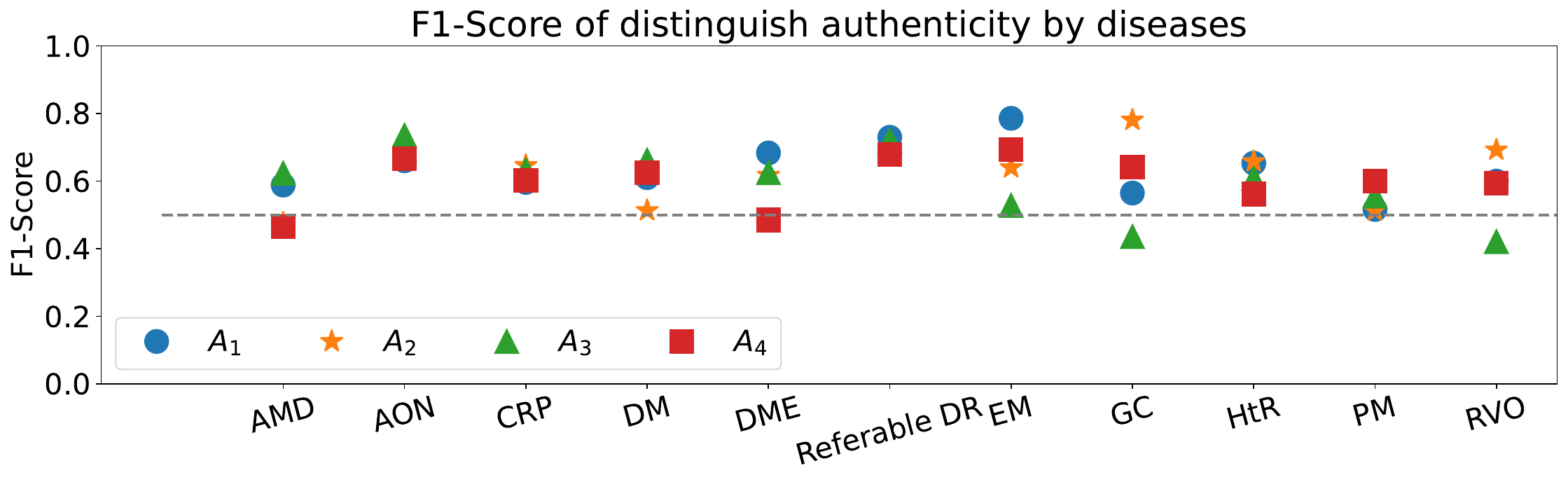}
\caption{\label{fig:human_f1}Distribution of F1-Scores by disease category for the four annotators' evaluations, with the gray dotted line representing the baseline F1-Score of 0.5 for random guessing. Scores significantly deviate from 0.5 suggest easier identification of the image's authenticity.}
\end{figure}

Furthermore, Figure~\ref{fig:human_f1} illustrates the distribution of F1-scores across different disease categories, revealing that the majority of F1-scores for various diseases fall within the range of 0.5 to 0.6, demonstrating the authenticity of synthetic images in various diseases. 
Notably, in some diseases, most annotators achieve relatively higher F1-Scores, such as referable DR (DR Grade $\geq$ 2) and GC, indicating a relative ease in authenticating these disease-specific images.
However, despite these synthetic images with some specific diseases can be easily distinguished by experienced annotators, experiments in Section \ref{sec:downstream_exp} and \ref{sec:pretrain} demonstrate that SynFundus-1M can still enhance the performance of models during pre-training or down-stream training w.r.t. DR and GC. 

\begin{figure}[!ht]
\centering
\includegraphics[width=1.0\columnwidth,trim=0 0 0 0,clip]{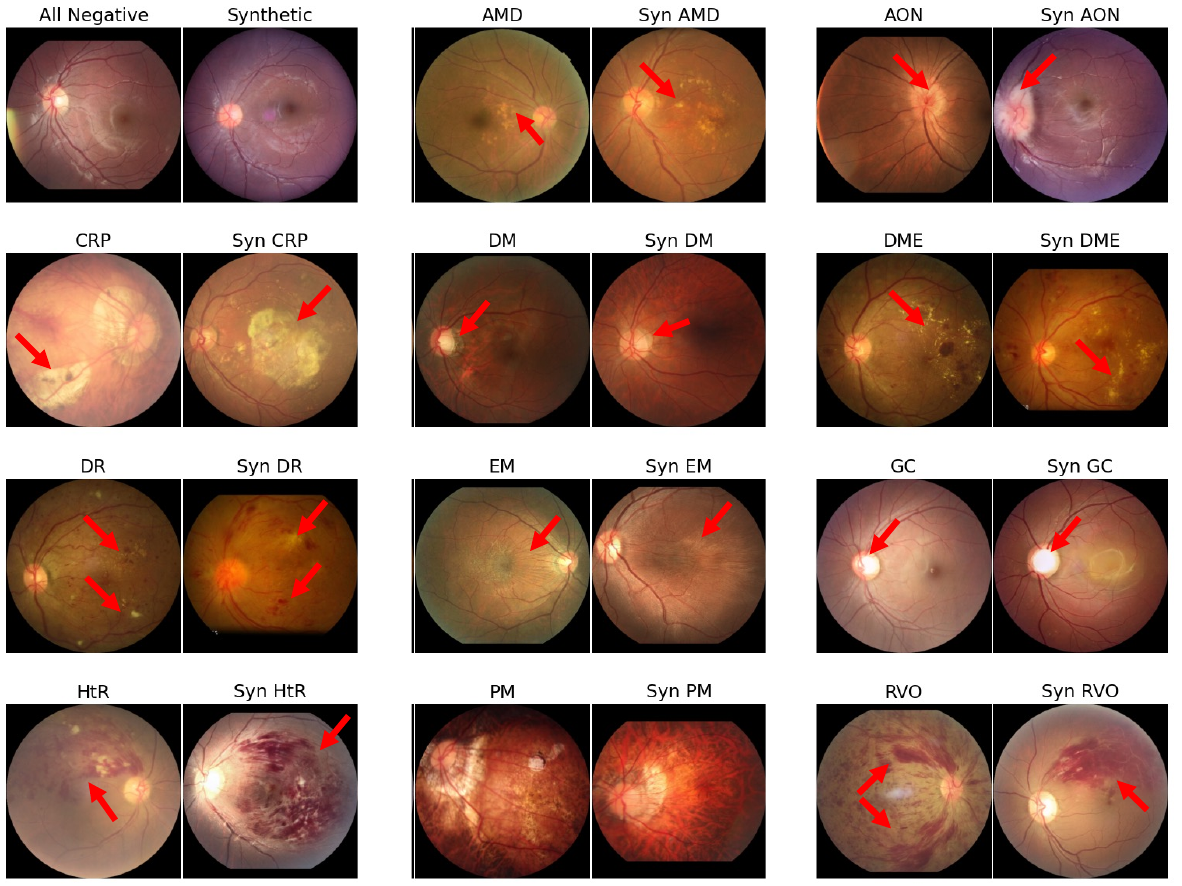}
\caption{\label{fig:visualize}Qualitative image generation comparisons. 
Authentic images (odd columns) are paired with SynFundus images (even columns) in various disease conditions. The disease-related visual features are annotated by red arrow. The fundus structure and lesions are very similar between authentic and synthetic images.}
\end{figure}

Additionally, Figure \ref{fig:visualize} presents a comparative examples between the synthetic images and authentic fundus photographs in terms of different diseases. We can see that the key visual features and lesions of the specific diseases are well generated in the synthetic images.

We utilize the Fréchet Inception Distance (FID) metric to evaluate the authenticity of the synthetic SynFundus-1M dataset, comparing to the existing authentic datasets and images generated using the MedFusion model. 
The results are listed in Table \ref{tab:fid}.
Lower FID scores indicate higher resemblance between the synthetic and authentic images.
To ensure a fair comparison, we standardize the calculation of FID between two datasets by randomly sampling a quantity of images equivalent to the smaller dataset from the larger one.

\begin{table}
\centering
\resizebox{0.6\linewidth}{!}{
\begin{tabular}{ccccc}
\hline
\multirow{2}{*}{Dataset} & \multirow{2}{*}{\#Images} & \multicolumn{3}{c}{FID $\downarrow $} \\ \cline{3-5} 
 &  & APTOS & EyePACS & AIROGS \\ \hline
APTOS \cite{aptos2019-blindness-detection} & 5,590 & - & 33.6885 & 32.5680  \\
EyePACS \cite{cuadros2009eyepacs} & 88,702 & 33.6885 & - & 7.5486*  \\
AIROGS \cite{devente23airogs} & 101,442 & \textbf{32.5680} & 7.5486* & -  \\ \hline
MedFusion \cite{muller2022diffusion} & 100,122 & 90.0301 & 63.0477 & 49.8271  \\
SynFundus-1M & 1,000,018 & 46.2985 & \textbf{29.7515} & \textbf{26.3920}  \\ \hline
\end{tabular}}
\caption{\label{tab:fid}FID (Fréchet Inception Distance) scores for various authentic and synthetic fundus image datasets. 
The table shows the FID scores when comparing each dataset against others. 
Lower scores indicate better resemblance. 
The best scores among the authentic and synthetic datasets are marked in bold. 
Note that EyePACS and AIROGS are selected from the same institution, and therefore, their internal comparison score (marked with *) is not considered when determining the best score.}
\end{table}

Table \ref{tab:fid} shows that SynFundus-1M's FID score closely matches those of authentic datasets, indicating that the discrepancy between SynFundus-1M and authentic datasets is akin to the variance observed among the authentic datasets themselves. 
Notably, since the EyePACS and AIROGS datasets are selected from the same institution, these two results lead to an exceptionally low and anomalous FID score of 7.5486. 
We further evaluated a dataset contains 100,122 images generated by the publicly available MedFusion model \cite{wu2023datasetdm}, and SynFundus-1M consistently achieve superior FID scores compared with MedFusion. 


\subsection{SynFundus for downstream disease diagnosis} \label{sec:downstream_exp}

\begin{table}
\centering
\resizebox{\linewidth}{!}{
\begin{tabular}{cccccccc}
\hline
\multicolumn{3}{c}{Training Set} & \multirow{2}{*}{Eval Set} & \multirow{2}{*}{Model} & \multicolumn{3}{c}{Metrics} \\ \cline{1-3} \cline{6-8} 
Main Data & Extra Data & \#Samples &  &  & QWK $\uparrow$ & Acc. $\uparrow$ & F1-Score $\uparrow$ \\ \hline
\multirow{9}{*}{IDRiD-Train} & \multirow{2}{*}{N/A} & \multirow{2}{*}{413} & \multirow{8}{*}{IDRiD-Eval} & ResNet-50 & 0.5590 & 0.5534 & 0.4448 \\
 &  &  &  & ViT-B/16 & 0.6190 & 0.4563 & 0.3499 \\
 & \multirow{2}{*}{EyePACS} & \multirow{2}{*}{89115} &  & ResNet-50 & 0.7503 & 0.6408 & 0.5123 \\
 &  &  &  & ViT-B/16 & 0.7764 & 0.6311 & 0.5093 \\
 & \multirow{2}{*}{SynFundus-1M} & \multirow{2}{*}{39769} &  & ResNet-50 & 0.7440 & 0.6602 & 0.5166 \\
 &  &  &  & ViT-B/16 & 0.7346 & 0.7087 & 0.6799 \\
 & \multirow{2}{*}{\begin{tabular}[c]{@{}c@{}}EyePACS\\ SynFundus-1M\end{tabular}} & \multirow{2}{*}{128502} &  & ResNet-50 & \textbf{0.8503} & \textbf{0.8058} & \textbf{0.7760} \\
 &  &  &  & ViT-B/16 & 0.8478 & 0.7864 & 0.7599 \\ \cline{2-8} 
 & \multicolumn{4}{c}{\begin{tabular}[c]{@{}c@{}}LzyUNCC Solution (IDRiD Top-1)\end{tabular}} & - & 0.7476 & - \\ \hline
\end{tabular}}
\caption{Performance evaluation of diabetic retinopathy grading on the IDRiD Dataset. 
The column labeled \textit{\#Samples} indicates the number of samples used for model training, suggesting that an increased size of the training set correlates with improved performance. 
Performance metrics indicate similar enhancements with the addition of equivalent scale of either authentic or synthetic images. 
Optimal results are achieved through the integration of both authentic and synthetic datasets, which is better than the Top-1 solution on the IDRiD benchmark.}
\label{tab:imagenet_idrid}
\end{table}

To assess the effectiveness of the SynFundus-1M for downstream tasks, we conducted experiments on two prevalent fundus imaging analysis topics: diabetic retinopathy grading and glaucoma diagnosis. According to Figure~\ref{fig:human_f1}, the synthetic images with these two diseases are relatively easy to be distinguished.

For disease classification, we select two representative architectures: ResNet-50\cite{he2016deep} representing traditional convolutional neural networks (CNNs), and ViT-B/16\cite{dosovitskiy2020image} exemplifying vision transformers (ViTs).

All models are initialized with the weights pre-trained on the ImageNet dataset and subsequently finetuned on different downstream datasets for 100 epochs. 
The input image resolutions differed between the models: ResNet-50 utilized images of 512x512 pixels, while the ViT-B/16 operated on a 384x384 pixel resolution. 
For optimization, ResNet-50 employed the Stochastic Gradient Descent with Momentum (SGDM) paired with a cosine decay learning rate strategy, with maximum rate of 1e-3. 
The ViT-B model is optimized using the AdamW optimizer with a constant learning rate of 1e-4. 
In order to provide a fair comparison with the Top-1 solutions and avoid potential data leakage caused by visible evaluation sets, only the evaluation metrics at the \textbf{last epoch} are listed.

\begin{table}[!ht]
\centering
\resizebox{\linewidth}{!}{
    \begin{tabular}{cccccccc}
    \hline
    \multicolumn{3}{c}{Training Set} & \multirow{2}{*}{Eval Set} & \multirow{2}{*}{Model} & \multicolumn{3}{c}{Metrics} \\ \cline{6-8} \cline{1-3} 
    Main Data & Extra Data & \#Samples &  &  & Acc. $\uparrow$ & AuC $\uparrow$ & F1-Score $\uparrow$ \\ \hline
    \multirow{9}{*}{REFUGE2-Train} & \multirow{2}{*}{N/A} & \multirow{2}{*}{1600} & \multirow{8}{*}{REFUGE2-Test} & ResNet-50 & 0.8075 & 0.8328 & 0.7333 \\
     &  &  &  & ViT-B/16 & 0.7450 & 0.8659 & 0.6919 \\
     & \multirow{2}{*}{AIROGS} & \multirow{2}{*}{103042} &  & ResNet-50 & 0.8950 & 0.8347 & 0.8047 \\
     &  &  &  & ViT-B/16 & 0.8850 & 0.8211 & 0.7962 \\
     & \multirow{2}{*}{SynFundus-1M} & \multirow{2}{*}{161600} &  & ResNet-50 & 0.8800 & 0.8595 & 0.8010 \\
     &  &  &  & ViT-B/16 & 0.8825 & 0.8432 & 0.7772 \\
     & \multirow{2}{*}{\begin{tabular}[c]{@{}c@{}}AIROGS\\ SynFundus-1M\end{tabular}} & \multirow{2}{*}{263042} &  & ResNet-50 & \textbf{0.9275} & 0.8761 & \textbf{0.8792} \\
     &  &  &  & ViT-B/16 & 0.9200 & \textbf{0.9150} & 0.8566 \\ \cline{2-8} 
     & \multicolumn{4}{c}{\begin{tabular}[c]{@{}c@{}}VUNO EYE TEAM Solution \\ REFUGE2 Final Match Top-1\end{tabular}} & - & 0.8827 & - \\ \hline
    \end{tabular}
}
\caption{Evaluating Performance in glaucoma diagnosis with the REFUGE2 Dataset. The column marked \textit{\#Samples} denotes the quantity of samples used in model training, highlighting the correlation between larger datasets and enhanced performance. 
The analysis demonstrates that adding an equivalent volume of either authentic or synthetic images yields similar performance improvements. 
The most significant enhancement in metrics is observed when the original REFUGE2-Train dataset is expanded with both AIROGS and SynFundus-1M datasets.}
\label{tab:imagenet_refuge2}
\end{table}

Table \ref{tab:imagenet_idrid} and Table \ref{tab:imagenet_refuge2} reveals two observations: 1) According to the relationship between \textit{\#Samples} and \textit{Metrics}, increased training data, whether synthetic or authentic, results in improved performance. 2) 
According to the second and third line from the top of Table \ref{tab:imagenet_idrid} and Table \ref{tab:imagenet_refuge2}, the evaluation metrics are comparable when added the same scale of extra data with authentic images(EyePACS and AIROGS) and SynFundus. 
Meaning that our synthetic images can expand the training data and improve the performance of downstream model as the authentic ones.
Moreover, compare to the performance in the last line of Table \ref{tab:imagenet_idrid} and Table \ref{tab:imagenet_refuge2}, even utilizing naive training strategies and model architectures, the model train with SynFundus-1M excelled beyond the top-tier solutions from the challenges.

Among the disease labels in SynFundus-1M, the scales of open-source datasets are rare expect DR and glaucoma. The experiments underscore the potential of the SynFundus-1M dataset in enhancing AI-driven diagnostics, particularly beneficial for the rare fundus diseases lacking open-source annotations.
 
\subsection{SynFundus for pretrain}
\label{sec:pretrain}

In this section, we pretrain ViT-B/16 and ResNet-50 models with SynFundus-1M to show that fundus models benefit from the released dataset in both performance and convergence speed.

\begin{table}
\centering
\resizebox{0.8\linewidth}{!}{
\begin{tabular}{ccccccc} 
\hline
\multirow{2}{*}{Dataset} & \multirow{2}{*}{Model} & \multirow{2}{*}{Pretrain} & \multicolumn{4}{c}{Metrics} \\
 &  &  & QWK $\uparrow$ & Acc. $\uparrow$ & AuC $\uparrow$ & F1-Score $\uparrow$ \\ 
\hline
\multirow{6}{*}{IDRiD} & \multirow{3}{*}{ViT-B/16} & Scratch & 0.0064 & 0.3010 & - & 0.1099 \\
 &  & ImageNet & -0.0063 & 0.3010 & - & 0.1026 \\
 &  & SynFundus-1M & 0.6607 & 0.6408 & - & 0.5103 \\
 & \multirow{3}{*}{ResNet-50} & Scratch & -0.0204 & 0.3010 & - & 0.1226 \\
 &  & ImageNet & 0.5752 & 0.4563 & - & 0.3928 \\
 &  & SynFundus-1M & \textbf{0.7412} & \textbf{0.6602} & - & \textbf{0.5241} \\ 
\cline{2-7}
\multirow{6}{*}{REFUGE-2} & \multirow{3}{*}{ViT-B/16} & Scratch & - & 0.8000 & 0.6651 & 0.4444 \\
 &  & ImageNet & - & 0.8000 & 0.7377 & 0.6783 \\
 &  & SynFundus-1M & - & 0.8450 & 0.8357 & 0.7664 \\
 & \multirow{3}{*}{ResNet-50} & Scratch & - & 0.8000 & 0.5065 & 0.4444 \\
 &  & ImageNet & - & 0.8250 & 0.8410 & 0.7491 \\
 &  &  SynFundus-1M & - & \textbf{0.8800} & \textbf{0.8645} & \textbf{0.7739} \\
\hline
\end{tabular}}

\caption{\label{tab:pretrain_metric}Comparison of fine-tuning performance using different pre-training strategies with ResNet-50 and ViT-B/16. 
Metrics encompass Quadratic Weighted Kappa (QWK), Accuracy (Acc.), Area Under Curve (AuC), and F1-Score. 
The best performances for each metric are highlighted in bold.}
\end{table}


\begin{figure}[!ht]
\centering
\includegraphics[width=0.9\columnwidth,trim=5 5 5 5,clip]{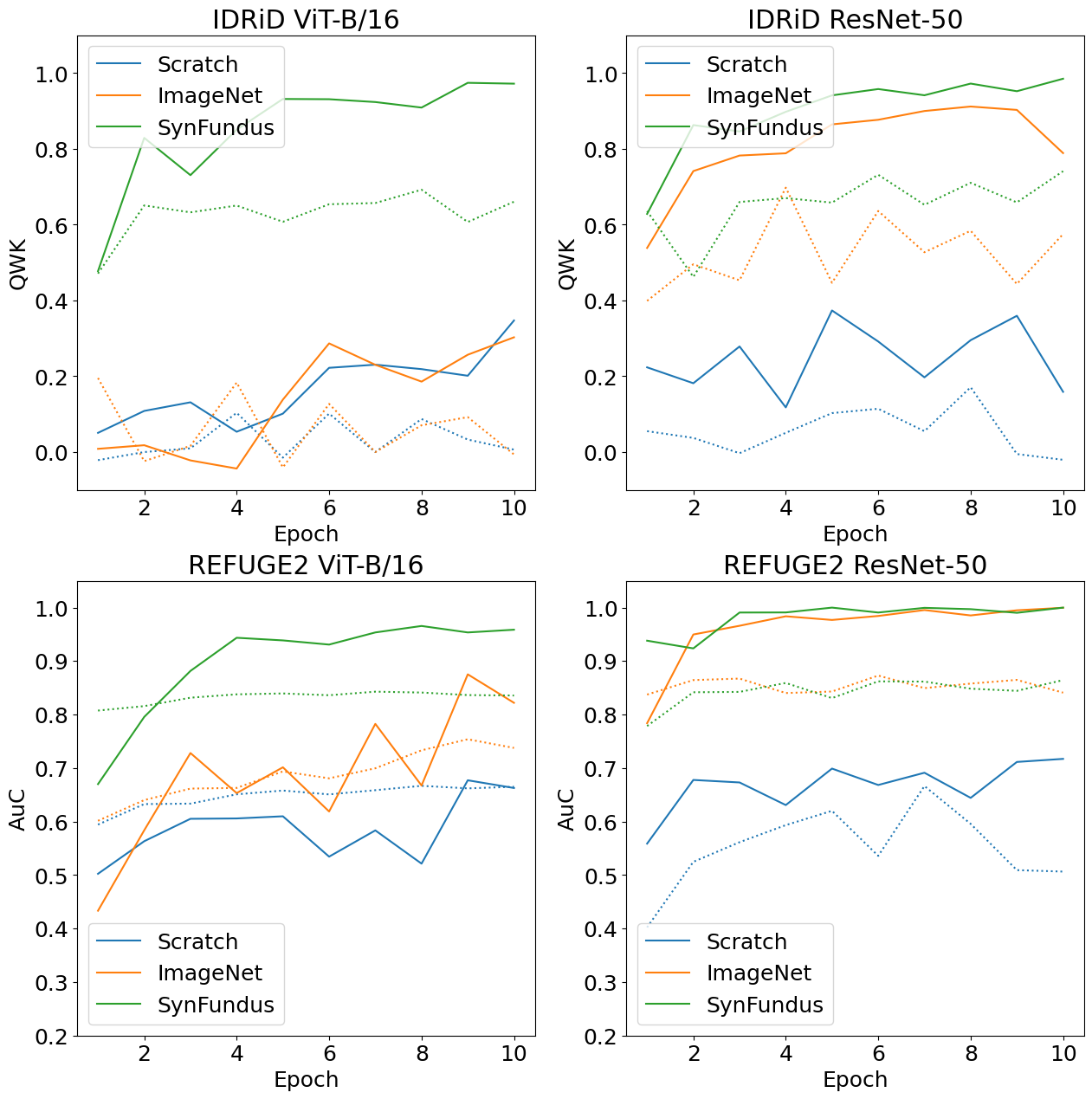}
\caption{\label{fig:convergence}Convergence trends of ViT-B/16 and ResNet-50 on IDRiD and REFUGE2, distinguished by polyline colors for pre-training configurations. 
Solid and dotted lines denote training and validation metrics respectively. 
Models pre-trained with SynFundus converge faster and achieves better final performance than ImageNets.}
\end{figure}

As illustrated in the Table \ref{tab:pretrain_metric}, models pretrained on SynFundus-1M consistently surpassed those pretrained on the widely used ImageNet dataset. Figure \ref{fig:convergence} illustrates the convergence curves of the finetuning processes for models pre-trained on SynFundus, ImageNet and from scratch. 
For both ViT-B/16 and ResNet-50 architectures, SynFundus-1M pre-training leads to rapid convergence on downstream tasks, which indicate that leveraging SynFundus-1M as pre-training data can substantially expedite the learning process and enhance overall model stability in downstream AI fundus tasks.

\section{Anonymization and privacy protection}


Recent research \cite{carlini2023extracting} has shown that publicly accessible diffusion models at the risk of leaking training data through specific generate-and-filter pipelines. For the SynFundus-1M synthetic images released to the public, to the best of our knowledge, no existing research has shown that specific semantic content in images, such as the direction of retinal blood vessels or the position of the optical disc, can be used to identify individuals precisely \cite{nakayama2023retinal}. 
Previous research has already demonstrated that synthetic fundus images can be effectively utilized for privacy-preserving training \cite{coyner2022synthetic} . Consequently, the release of SynFundus-1M can significantly benefit AI research in fundus imaging, while simultaneously mitigating the risk associated with data privacy.

Furthermore, we have meticulously executed the following protocols throughout this work:

\begin{itemize}
    \item \textbf{Raw Data Management}: The private datasets deployed in this research are securely ensconced within specially fortified servers, residing in security zone. 
    These servers are equipped with GPU and maintain a strict network isolation policy. 
    All model training tasks using this data must be completed within the server. 
    Large-scale data transmissions are strictly curtailed. 
    Even minor data transmissions, for instance, trained model parameters, necessitate explicit approvals and are shepherded out of the security zone by dedicated personnel.

    \item \textbf{Training Data of the SynFundus-Generator}: During the assembly of the generator's training dataset, we employed the Hough Circle Transform algorithm. 
    This algorithm identifies the largest continuous circular Region of Interest (ROI) in each image, which in turn omits the peripheral black areas commonly laden with text-based metadata. 
    This step virtually eradicates any textual residue that might inadvertently compromise privacy.

    \item \textbf{Further Checking of SynFundus-1M}: 
    To ensure the absence of identifiable text within SynFundus-1M, we employed the advanced PP-OCRv4 \cite{ppocr} for optical character detection, thereby confirming the absence of any identifiable characters in the images. 
    In addition to automatic evaluation, we also randomly selected 2,000 images from SynFundus-1M for manual inspection. 
    We found no identifiable characters in the images or any identifiable data linking them to a specific individual person.
\end{itemize}

\section{Discussion}

Due to privacy considerations, we are unable to share the distribution and detailed information of our private dataset. 

Although the synthetic images generated by SynFundus-Generator have shown promising results, the SynFundus-1M has limitations:

\textbf{The capacity of the generation model}: In this study, the SynFundus-Generator is a DDPM-based model which is not currently considered as the state-of-the-art among diffusion models. the future research could explore more advanced methodologies to generate fundus and medical images, such as DiT\cite{Peebles2022DiT}, to potentially enhance generative performance. 

\textbf{The noises in the annotated labels}: The automated annotations of the SynFundus-1M are produced by the models in our AI-diagnose platform, potentially being less than optimal due to model capacity constraints.
In practical evaluations, the annotation models exhibit approximately 90\% sensitivity and specificity, with a potential 10\% margin attributed to noise. 
Researchers are encouraged to refine the annotations of the SynFundus-1M as needed or utilize their own diagnostic models to boost annotation precision.

\textbf{The content tendencies in synthetic images}: SynFundus tends to produce synthetic images that depict exaggerated symptoms of diseases, such as widespread hemorrhaging or indistinct optic disc boundaries. 
Since these severe symptom expressions constitute distinct visual features, their conspicuous nature allows for more effective identification and assimilation by the generative model. As a result, most of the disease data in the dataset tends to be typical, while there is a relative scarcity of data with long-tail disease features.

\section{Conclusion}

To fulfill the community's need for extensive, high-quality annotations on large-scale fundus datasets, we introduce SynFundus-1M, a comprehensive set of synthetic fundus images with extensive annotations.
The synthetic dataset exhibits remarkable resemblance to authentic fundus images, especially in diseases featuring prominent and extensive lesions. 
We have proved the value of SynFundus-1M through comprehensive experimentation, demonstrating that the mainstream visual architectures (both CNNs and ViTs) can enhance their performance in disease diagnosis tasks through pre-training or fine-tuning. We hope that the SynFundus-1M can promote the development of sophisticated models in fundus disease analysis.

\section{Acknowledgment}

We are grateful for the contributions made by Jia Liu, Dalu Yang, and Wenshuo Zhou in conducting experiments to distinguish the authenticity of SynFundus-1M. 
We also extend our heartfelt thanks to Professor Yanwu Xu, Lei Wang, Qinpei Sun, Binghong Wu, Xu Sun, and Tiantian Huang for their outstanding work in developing the AI-diagnosis platform during their tenure at Baidu. Furthermore, our collaborating ophthalmologists and annotators have also made valuable contributions to this study by annotating diseases in the private dataset.

%
%
\bibliographystyle{splncs04}
\bibliography{main}

\end{document}